# A Unified Quantitative Model of Vision and Audition

Peilei Liu[†], Ting Wang[†]

**Abstract**: We have put forwards a unified quantitative framework of vision and audition, based on existing data and theories. According to this model, the retina is a feedforward network self-adaptive to inputs in a specific period. After fully grown, cells become specialized detectors based on statistics of stimulus history. This model has provided explanations for perception mechanisms of colour, shape, depth and motion. Moreover, based on this ground we have put forwards a bold conjecture that single ear can detect sound's direction. This is complementary to existing theories and has provided better explanations for sound localization.

Great strides have been made in vision and audition research[1-6]. Nevertheless, there remain substantial gaps between physiological evidences and existing models[6-11]. We wish to put forwards a unified quantitative model on the cell scale. It has provided explanations for possible biological perception mechanisms of shape[8], colour[9], motion[10], depth[11] and sound localization[5, 6]. Specifically, this is a self-organized model composed of cell model and synapse model. All $c_i$ in this paper are constants, and they have different meanings in different paragraphs.

**Cell model (A)**: A1) for all cells: $v = c_1(1-e^{-c_2\sigma})$, $\sigma = \sum w_i v_i$, where v and $v_i$ are the output and input graded potentials (or spike frequencies) respectively, $w_i$ is the synaptic strength. A1 is inspired by MP model[12] and Hodgkin-Huxley model[13]. A2) for horizontal and amacrine cells: $\frac{dv_i}{dt} = -h_v$, $h_v = c_3(1-e^{-c_4\sigma'})$, $\sigma' = \sum_{j \neq i} v_j$, where $h_v$ is the hyperpolarized potential of output. Namely A2 will result in lateral inhibition through electrical synapses, which is known as "MAX" operation or "winner-take-all" competition in existing models[8, 14]. For horizontal cells $t \leq t_0$, namely the competition is "soft". A3) for photoreceptors: $v = c_5 e^{-c_6 d}$ $(d \leq m_d)$, where d means the distance between the cell and stimulus associated with specific attribute. For example, for colour d means distance between wavelengths. Since horizontal and amacrine cells are subsidiary, bipolar and ganglion cells are called cells for short.

**Synapse model**: A4) $w_i = c_1(1-e^{-c_2 r_i})$, $\frac{dr_i}{dt} = c_3(c_4 - \sum r_i)F(v_i) > 0$, where the specific form of function $F(v_i)$ will be given later. And $r_i$ should be a kind of limited synaptic resource. Activated cells tend to allocate this resource according to the stimulus pattern $v_i$. A4 is inspired by the Hebb conjecture[15] and BCM model[16]. A5) $\frac{dr_i}{dt} = -r_7 w_i$ when $v_i$=0. Namely synapses decay and release resource r when without stimulus. And the decay rate is viewed as constant $c_7$ here for simplicity. This synapse model is mainly about dendritic synapses of bipolar and ganglion cells, while synaptic strengths of horizontal and amacrine cells are constants. According to this model, synapses are alive and self-adaptive like muscle fibers: Exercises make them grow, but they will decay passively without exercises. However, they are plastic only in a critical period[17, 18], after which they become stable. This should be the physiological foundation of monocular deprivation. Specifically, synapses in the closed eye will decay and release the synaptic resource r, which will be taken away by the open eye. As well known, nerve cells without succeeding

[†]College of Computer, National University of Defense Technology, 410073 Changsha, Hunan, China

connections will die[1], and then vision will be deprived permanently. Similarly, the cortical visual areas of the blind will be taken over by auditory afferent fibers.

The cell model and synapse model are significant in statistics and computation. Suppose the probability of an object occurring on condition of attribute $A_i$ is $P(O_i)= P(O|A_i)$, where $A_i$ are independent events. Let $P(A_i)=v_i/c_1$, $P(\neg O_i)= e^{-c_2 v_i}$, and then $P(O)=P(O_1+O_2+\ldots+ O_n)= 1-P(\neg O_1 \neg O_2 \ldots \neg O_n)=(1-e^{-c_2\sigma})$, where $\sigma = \sum_{i=1}^{n} v_i$. Compared with A1, the neuron's output potential $v=c_3 P(O)$ can be viewed as the probabilistic estimate of an object's occurring, while the input $v_i$ can be viewed as the probability of attributes occurring $P(A_i)$. Similarly, $w_i = c_4(1-e^{-c_5 r_i})$ in A4 could be viewed as the statistical confidence of an attribute on condition of stimuli history. In essence, A4 and A1 reflect temporal and spatial localization respectively. Namely object tend to exist in continuous local space-time. In conclusion, a cell is a statistical machine in essence, consistent with the conjecture of John Von Neumann[19] and the statistical physics[20]. From the computational viewpoint, the gradually slower growth of synapses in A4 favors neural network converging to global minimum point, similar to the simulated annealing algorithm[21]. Incidentally, main constants in both models are exponential functions which could be implemented easily in biology and physics. For instance, radioactive materials decay exponentially.

Definition 1: a detector c of an input I means: $f_c(I)=1$, while $f_{c'}(I)=0$ for all $c'\neq c$. Therefore a detector is a well-defined function in essence. Namely there exists a "mult-to-one" mapping from inputs to detectors, while "one-to-mult" correspondence and replicative coding could be prevented by lateral inhibition. In brief, cells are specialized in the work division.

Theorem 1: a bipolar cell can detect the colour of a point. As shown in Fig. 1B, the colour of a point stimulus is between two cones, with wavelength distances from them $d_1$ and $d_2$ respectively, and $d_1 + d_2 = m_d$. For example, orange is between red and green. According to A3, their graded potentials $v_i = c_1 e^{-c_2 d_i}$ (i=1, 2). The bipolar cell activated most will inhibit its neighbors and enlarge the potential disparities through horizontal cells according to A2. And its dendritic synapses will be strengthened according to A4, namely $w_i = c_3(1-e^{-c_4 r_i})$. According to A1, $\sigma = \sum_{i=1}^{2} w_i v_i = c_5 \sum_{i=1}^{2} e^{-c_2 d_i}(1-e^{-c_4 r_i})$. It can be inferred that the output v and σ get the maximum values when $r_i = c_6(m_d - d_i) = c_7 \log(v_i) + c_8$. Namely $F(v_i)$ is linear with $\log(v_i)$ in A4. Therefore an activated cell actually strengthens its dendritic synapses towards the potential pattern, which will bring competitive advantage reversely. When it becomes the unique winner in the "winner-take-all" competition, it is the detector of this colour according to definition 1. And its rivals could be viewed as background noise. With the stimulus continuing, this detector will become more precise. In most cases however, a detector should be a statistical pattern of many similar inputs. And a new input is actually merged into the most similar detector cell. Gaussian function is often used as the spatial decay curve of photoreceptor's potential in A3[8]. In this case however, the relation between $r_i$ and $d_i$ isn't linear any more. Moreover, exponential function can be more easily implemented in biology.

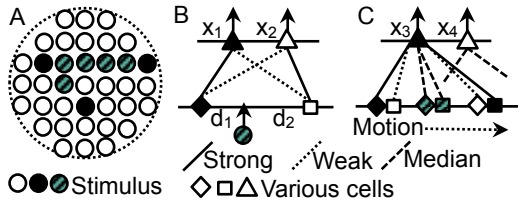

**Fig. 1.** Detection of point's attributes and spatial frequency. The gray levels of points mean intensities of input attributes or potentials. As in A, every input is a spatial graded pattern of attributes in essence. The attribute could be any type such as color, luminance, or time delay. For example, motion could be a spatial delay pattern of points as in C. A grey point in A can be transformed to the potential pattern of two cells in B. And then it can be detected by a single cell with specific synaptic strength pattern such as $x_1$. As in C, the spatial frequency of points can be detected by a cell with proper synaptic strength pattern such as $x_3$. In fact, $x_3$ can detect a whole input within its crown completely. Generally speaking, the synaptic strength should be positive correlative with the presynaptic potential. Incidentally, lateral inhibition is indispensable in this detection.

Corollary 1: a cell can detect a point's luminance, time delay, and spatial position. In essence, colour is a point in the light spectrum. Similarly, the luminance is a point in the luminosity space, while a delay is a point in a time stream. As long as d in A3 means luminance disparities, time intervals and spatial distance, a ganglion cell can detect luminance, time delay and spatial position respectively (see Fig. 1B). To prevent mingling, these physical quantities are detected by different layers of cells in retinas. Specifically, photoreceptors detect position; bipolar cells detect colour; ganglion cells detect luminance and time delay. Moreover, the central retina detects physical quantities different from around as well. And these detections need the assist of specific sensory cells similar to cones. Specifically, luminance detection may need special bipolar cells responding to hyperpolarized potentials. And this is supported by the two types of receptive fields "on" and "off"[22]. Delay detection however needs special ganglion cells sensitive to time delay, namely their dendritic synapses have constant time delays. Specially, every rod cell has a specific spatial position. Since rods are arranged compactly namely $d \approx 0$ in A3, each rod is a natural position detector. In contrast, there are only three kinds of cones in the broad spectrum space, which is very sparse. In essence, spatial length and time delay are ordinary physical quantities, similar to colour and luminance. And all attributes of single point can be detected similarly.

Theorem 2: a ganglion cell can detect spatial frequency of points. Suppose that n bipolar cells are activated within a limited range (see Fig. 1A) and the ganglion cell activated most has m dendritic connections. Similar as in theorem 1, it will inhibit neighbors through amacrine cells according to A2, and strengthen dendritic synapses according to A4 $w_i = c_1(1-e^{-c_2 r_i})$. When $m \geq n$, $\sigma = \sum_{i=1}^{n} w_i v_i = c_3 \sum_{i=1}^{n}(1-e^{-c_2 r_i})v_i$ according to A1. Since $\sum_{i=1}^{m} r_i = c_4$ according to A4, $\sigma$ gets the maximum value when m=n. When $m \leq n$, $\sigma = \sum_{i=1}^{m} w_i v_i = c_3 \sum_{i=1}^{m}(1-e^{-c_2 r_i})v_i$. Since $1-e^{-nx/m} \leq (n/m)(1-e^{-x})$, $\sigma$ gets the maximum value when m=n. In conclusion, v in A1 gets the maximum value when m=n. Therefore, a perfect detector should connect to all points with proper synaptic strength pattern according to definition 1. In most cases however, this is only a best approximation rather than precise match, and a detector can actually detect many similar spatial frequencies. In fact, there could be two kinds of ganglion cells with different receptive fields. According to A2, $v_i' \approx v_i - h_v = c_4 e^{-c_5 n} + c_6$ when ignoring the differences of $v_i$. And then in A1 $\sigma = \sum_{i=1}^{n} w_i v_i = c_7 + c_8 \sum_{i=1}^{n}(1-e^{-c_2/n})(e^{-c_5 n}+c_9)$. This function has maximum value at a specific point $n=n_0$. Specifically, when $c_2 \ll c_5$, $n_0 \approx 1$. The ganglion cell is more sensitive to single point in this case, namely with the receptive field of concentric circles[22]. Otherwise when $c_2 \gg c_5$, $n_0$ is very large. And ganglion cell is more sensitive to background such as full light.

Corollary 2: cells can detect any input within their crowns. Any input is a spatial pattern of attributes intensities (see Fig. 1A). The attributes of single point can be detected according to theorem 1 and corollary 1. And the spatial frequency of points can be detected according to theorem 2. In fact, both of them can be detected meanwhile by a single cell with proper synaptic strengths pattern (see Fig. 1C). And the specific form of $F(v_i)$ in A4 is determined by the constraints between $v_i$. Although a ganglion cell can detect the whole input completely in theory, the nerve system actually uses a hierarchical network instead[8], namely a ganglion cell only detects a small part of an input. The benefit is that the detection and encoding are fine-grain, and resource can be saved through sharing common median nodes[23]. And the cost is slower speed.

In conclusion, nerve cells are computing units as well as storage units, whose functions are determined by their dendritic connections. They become specialized detectors in fully grown eyes. Therefore synapses are hardware encoding, while software is the graded potentials or firing frequency. In essence, the retina is a hierarchical feedforward network composed of special sensors. Interneurons however are universal sensors, and therefore circuits can form. Some important mechanisms can only be explained by circuits such as feature constancy and sequence encoding. Our model has actually reconciled "grandmother cells" and population coding[24]. In other words, population coding is transformed to "grandmother cell" through the retina. As results, input to the cortex is sparse, which is digital coding in essence. The "grandmother cell" is questioned usually because neurons are too few for coding the infinite objects. According to our model however, a cell can coding more than one object. And it isn't a template of all stimuli in its receptive field. Instead, it mainly extracts features in contrast to background noise through space-time adaptive mechanisms, such as lateral inhibition and fatigue of photoreceptors. As computing units, neurons themselves instead of brain rhythm should be the "binding mechanism" in population coding[1].

**Motion perception**. Since displacement is the product of velocity and time: s=vt, velocity of motion is determined by time delay when displacement is constant[10]. Therefore a motion is a spatial pattern of time delay in essence (see Fig. 1C), which can be detected by a cell according to corollary 2. From this viewpoint, motion perception is the same as coincidence detection in the auditory research[5]. Therefore, motion is actually equivalent to a sequence of pictures for the nerve system. That might be why we can percept motion in movie. It is interesting that a motion can be encoded by a static cell. In fact, static objects can be viewed as motions without latency, namely velocity is infinite. That might be why computer screen and propeller rotating fast seem static.

**Depth perception.** Depth perception is the visual ability to perceive the distance of an object[4]. We suppose that the vision system only knows information about itself, such as the position of focal point which could be gained through oculomotor nerves. In this case, almost all monocular cues can be excluded, such as relative size and motion parallax. In order to use these monocular cues, the vision system has to know the real size or motive speed. However, this requires knowing the distance reversely. This is actually a vicious circle. As shown in Fig. 2, depth perception can be transformed to the disparity measurement problem[4, 11]. According to our model however, it doesn't even need to compute disparities: It just simply overlaps two visual fields and save all features in the spatial pattern to a single cell. Then this cell can detect depth according to corollary 2. In my opinion, some existing theories mix up depth perception and depth illusion caused by visual experience. Any 2D picture will be matched with existing 3D detectors in our visual system. And some deceiving cues will make them match well, such as shadow and masking. In this case, depth illusion will be caused. In the famous Müller-Lyer illusion for example[25], if a line segment seems far away according to existing clues, it will be inferred that it should be longer when nearby. This inference process could be conducted in cortical circuits, similar to feature constancy. And it is right in the real 3D world at most cases. Therefore these man-made 2D pictures other than our eyes deceive us.

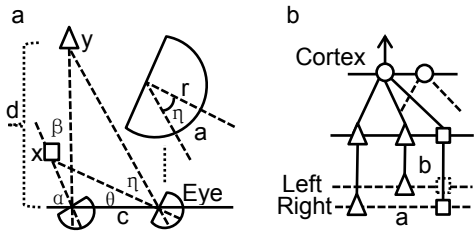

**Fig. 2.** Depth perception. In A, x is the focus point of both eyes. Positions of points x and y in the retinas are shown in B. The vertical distance between x and y can be computed precisely as following: $d = c \times \tan(\alpha+\beta)\tan(\theta+\eta)/[\tan(\alpha+\beta)-\tan(\theta+\eta)]$, where angles $\alpha$ and $\theta$ could be gained from oculomotor cues. And angles $\beta$ and $\eta$ could be computed through disparities: $\beta = b/r$, $\eta = a/r$, where r is the radius of eyeball. As in B, if we simply overlap the two retinas, these disparities can be viewed as a spatial pattern in Fig. 1A.

**Sound localization**. According to existing theories, sound localization is through interaural differences[6, 26]. However this is difficult in the median plane (front-above-back-below) or the cone of confusion[27]. In addition, ultrasonic used by bats is a disadvantage for the detection of interaural time differences[28]. Spectral cues such as pinna effect and head shadow are supposed to help[26, 29]. However, this is equivocal for animals with small pinnae and heads such as bats and flies. We put forward a bold conjecture that a single ear can detect the incidence angle of sound (see Fig. 3). Similar to light, sound is a kind of wave propagating rectilinearly somewhat. Therefore its incidence angle corresponds to a specific position in cross-section of cochlea, which could be detected according to corollary 1. Evidences from monaurally deaf listeners actually support this conjecture[29]. And it should be feasible in physiology as well. For example, voice frequency is actually detected by the depth of cochlea according to the traveling wave theory[30]. From this viewpoint, the cochlea is similar to retina. And sound location should be similar to the three-dimensional vision, namely through disparities in cochleae (see Fig. 2). Incidentally, this suggests a novel stereo earphone similar to the 3D glasses, namely with multiple sound sources distributed in the rim (see Fig. 3). According to our conjecture, the ultrasonic with better directionality is better at sound localization. This conjecture can also explain why the sound location becomes weak for sound sources far away similar to the visual depth perception. In fact, our conjecture is a complement to existing theories other than replacing them. Single ear is good at detecting median plane, while interaural differences between two ears are good for detecting left-right direction. Similarly, disparities and echo interval can also be combined by bats for detecting the distance of sound source. Different clues can be binded together by nerve cells.

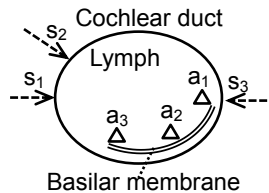

**Fig. 3.** Single ear detecting sound's direction. The cross-section of cochlea is shown. Sound $s_1$, $s_2$ and $s_3$ generate maximal stimulus at point $a_1$, $a_2$, $a_3$ in the basilar membrane respectively. Therefore their directions could be detected by nerve cells according to theorem 1. This suggests a novel earphone with three independent sound sources embed in the three positions a1, a2, a3 respectively. Then sound's incidence angles to both ears can reserve, and therefore the stereo effect could be improved like the 3D glasses.

Other sense such as tactile sense and olfactory sense could also follow similar coding mechanisms. Differently however, smell doesn't propagate rectilinearly. This may be why we have two ears but only one nose.

**Author Information** The authors declare no competing financial interests. Correspondence and requests for materials should be addressed to Peilei Liu (e-mail: plliu@nudt.edu.cn)